\documentclass{article}

\usepackage[preprint]{neurips_2022}

\usepackage[utf8]{inputenc} 
\usepackage[T1]{fontenc}    
\usepackage{hyperref}       
\usepackage{url}            
\usepackage{booktabs}       
\usepackage{amsfonts}       
\usepackage{nicefrac}       
\usepackage{microtype}      
\usepackage{xcolor}         
\usepackage{multirow}
\usepackage{graphicx}
\usepackage{amsmath}

\title{Delving into the Pre-training Paradigm of Monocular 3D Object Detection}

\author{%
  Zhuoling Li \\
  Tsinghua University\\
  \texttt{lzl20@mails.tsinghua.edu.cn} \\
  \And Chuanrui Zhang \\
  Tongji University \\ 
  \texttt{1853513@tongji.edu.cn} \\ 
  \And En Yu \\
  Huazhong University of Science and Technology \\
  \texttt{yuen@hust.edu.cn} \\
  \And Haoqian Wang\thanks{Corresponding author.} \\
  Tsinghua University \\
  \texttt{wanghaoqian@tsinghua.edu.cn}
}

\begin{document}

\maketitle

\begin{abstract}
The labels of monocular 3D object detection (M3OD) are expensive to obtain. Meanwhile, there usually exists numerous unlabeled data in practical applications, and pre-training is an efficient way of exploiting the knowledge in unlabeled data. However, the pre-training paradigm for M3OD is hardly studied. We aim to bridge this gap in this work. To this end, we first draw two observations: (1) The guideline of devising pre-training tasks is imitating the representation of the target task. (2) Combining depth estimation and 2D object detection is a promising M3OD pre-training baseline. Afterwards, following the guideline, we propose several strategies to further improve this baseline, which mainly include target guided semi-dense depth estimation, keypoint-aware 2D object detection, and class-level loss adjustment. Combining all the developed techniques, the obtained pre-training framework produces pre-trained backbones that improve M3OD performance significantly on both the KITTI-3D and nuScenes benchmarks. For example, by applying a DLA34 backbone to a naive center-based M3OD detector, the moderate ${\rm AP}_{3D}70$ score of Car on the KITTI-3D testing set is boosted by 18.71\% and the NDS score on the nuScenes validation set is improved by 40.41\% relatively.
\end{abstract}

\section{Introduction}
\label{Sec: Introduction}
\vspace{-0.2cm}

As a fundamental task of perceiving the 3D real world, monocular 3D object detection (M3OD) has been widely adopted in numerous practical applications, e.g., autonomous driving and robotic navigation \cite{liu2020smoke}. Due to the rapid development of deep learning \cite{he2016deep,lecun2015deep}, the state-of-the-art (SOTA) performance of M3OD is boosted dramatically in recent years. However, current deep learning technology still relies on numerous labeled data for training, and the process of annotating 3D bounding boxes for M3OD is expensive. Therefore, studying how to exploit the information contained in unlabeled data is meaningful.

To this end, many unsupervised pre-training algorithms are proposed, such as MoCo \cite{he2021masked} and MAE \cite{he2021masked}. Although these algorithms present promising performance on downstream tasks of 2D visual perception, our experimental results indicate that their effectiveness on M3OD is limited. Meanwhile, the only work we know about M3OD pre-training is DD3D \cite{park2021pseudo}, the pre-trained backbones of which have been confirmed effective and employed by some recent M3OD detectors \cite{li2022bevformer,liu2022petr}. However, the pre-training task adopted by DD3D is only direct depth estimation, and we find that its effectiveness may be attributed to the numerous extra private data (about 15M frames). When we conduct the same pre-training procedure on smaller data volume (such as pre-training on KITTI-raw and then evaluated on KITTI-3D \cite{geiger2012we}), the pre-training does not bring performance gain. This observation implies that direct depth estimation is inefficient for M3OD pre-training, and a more powerful paradigm is desired.

\begin{figure}[tbp]
    \centering
    \includegraphics[scale=0.34]{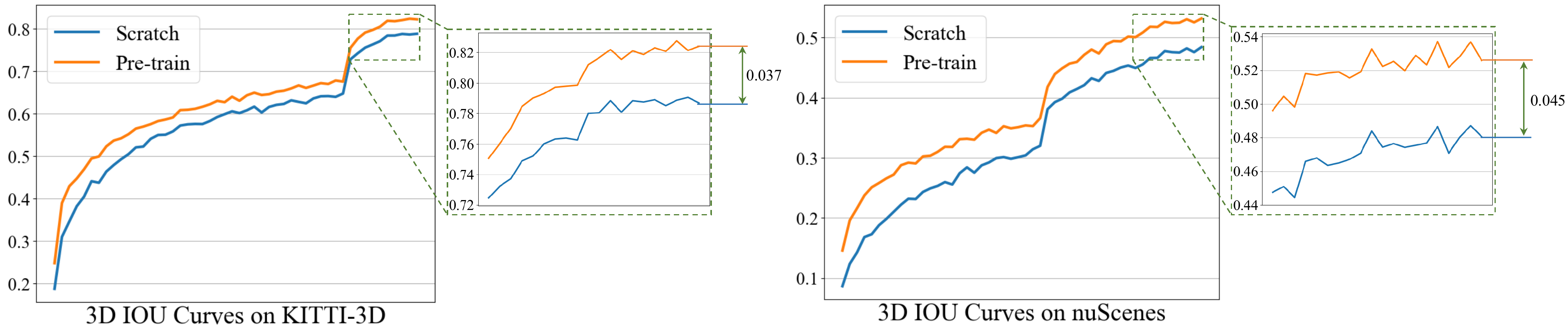}
    \vspace{-0.2cm}
    \caption{The influence of DEPT on the 3D IOU curves between predicted boxes and ground truth boxes during the training process.} \label{Fig: IOU curves}
    \vspace{-0.4cm}
\end{figure}

In this work, we aim to bridge this gap. Since there exist few previous works about M3OD pre-training, we first need to explore the guideline of devising pre-training tasks. Through careful analysis, we find that the key point is making the pre-training task representation imitate the target task representation. Besides, we argue that it is difficult to imitate the representation of M3OD if no any kind of labels is available, because unsupervised pre-training algorithms only produce image-level appearance representation \cite{li2022efficient}. Conversely, M3OD demands instance-level depth and appearance representation \cite{zou2021devil}. 

Meanwhile, we hope the used labels can be generated without manual effort. In this way, we are able to pre-train models with a cost similar to unsupervised pre-training. Through analysis, we select depth estimation to imitate the M3OD depth representation and 2D object detection to produce appearance representation. Among them, the depth labels come from lidar points and the 2D box labels can be obtained via an already well-trained 2D object detector. Our experiments reveal that the qualities of both them are reliable. Additionally, the experiments also suggest that combining depth estimation and 2D object detection is an efficient baseline.

Afterwards, we further delve into this baseline, and devise strategies to better imitate M3OD representation. For the depth estimation part, we only predict the depth in concerned areas and propose a strategy that augments the sparse depth labels into a semi-dense form. For the 2D object detection part, we observe that the sensibility of models to keypoints is important, and develop a strategy to improve this ability of pre-trained models. Furthermore, we design a class-level loss adjustment strategy to make the learned representation of various classes more balanced.

Combining all the techniques and insights, the derived M3OD pre-training framework, namely DEPT (the abbreviation of ``depth and detection''), produces promising pre-trained backbones. As illustrated in Fig. \ref{Fig: IOU curves},  the pre-trained backbones improve the performance of M3OD detectors on both KITTI-3D and nuScenes significantly. In addition, DEPT presents some extra benefits: (1) DEPT alleviates the long-tailed distribution problem \cite{zhang2021deep} in M3OD significantly. (2) The representation learned from DEPT is easy to generalize across various domains (such as between KITTI and nuScenes).

\vspace{-0.2cm}
\section{Related Work}
\label{Sec: Related Work}
\vspace{-0.2cm}

\paragraph{Monocular 3D Object Detection.} In contrast to 2D object detection, M3OD \cite{chen2016monocular,li2022diversity} is for recognizing objects of interest in monocular images and estimating their properties including location, dimension, orientation, etc. According to the form of predicted depth, existing M3OD methods can be divided into two groups, i.e., the dense-depth based and sparse-depth based methods.  The dense-depth methods need to predict a depth value for every pixel of the input image, the process of which is time-consuming \cite{ma2020rethinking,manhardt2019roi,shi2020distance}. Therefore, the inference speed of dense-depth based methods is often limited. However, the dense-depth methods often behave better in detecting small and non-rigid targets. By contrast, the sparse-depth based methods directly regress the 3D center locations of concerned objects \cite{li2021monocular}, which is more flexible and faster. Among the sparse-depth based methods, some use direct estimation \cite{liu2020smoke} and some others rely on geometric constraints to obtain depth \cite{li2020rtm3d}.

\paragraph{Pre-training Algorithms.} Pre-training models on large-scale data has been confirmed valuable by both the natural language processing and computer vision communities \cite{brown2020language,riquelme2021scaling}. The obtained models often present promising generalization ability and can be used to boost the performance of downstream tasks \cite{li2021enabling}. To satisfy the demand of different downstream tasks, various pre-training tasks are designed, such as contrastive learning (CL) \cite{chen2020simple}, masked autoencoders (MAE) \cite{he2021masked}, super resolution \cite{ledig2017photo}, pixel prediction \cite{chen2020generative}, etc. Nevertheless, almost all of them are for 2D visual perception. The publications about how to design pre-training tasks for 3D visual perception are scarce. The only work we know about M3OD pre-training is DD3D \cite{park2021pseudo}, which is only a direct depth estimation task. The guideline of how to devise efficient M3OD pre-training tasks is still unclear.

\vspace{-0.2cm}
\section{Guideline and Basic Paradigm}
\label{Sec: Guideline and Basic Paradigm}
\vspace{-0.2cm}

Since designing M3OD pre-training tasks is a hardly studied problem, we first explore the guideline of it. Through revisiting existing publications, we observe an interesting phenomenon: A pre-training task generally benefits the downstream tasks similar to itself. For example, CL is inherently a classification task treating the views augmented from the same image as the same class \cite{chen2021exploring}. The often adopted downstream tasks to demonstrate its superiority include classification, 2D object detection and segmentation, which are image-level, instance-level, pixel-level classification tasks, respectively. However, other types of tasks such as 3D reconstruction usually do not employ the models pre-trained by CL. Similarily, CLIP \cite{radford2021learning} trains a model to capture the relation between images and texts, and ViLD \cite{gu2021open} successfully improves the performance of open-vocabulory object detection with the pre-trained model of CLIP. Based on this observation and the insights in the InfoMin principle \cite{tian2020makes,tishby2000information}, we speculate that the guideline of designing pre-training tasks is \textbf{making the pre-training task representation imitate the downstream target task representation}. Thus, we need to design a pre-training task imitating the representation of M3OD.

Suggested by \cite{zou2021devil}, the representation of M3OD can be divided into two parts, the depth representation and appearance representation. To verify the aforementioned guideline, we analyze how depth estimation pre-training affects the various loss items in M3OD training process. It is found that the convergence of depth related losses is acclerated significantly, while the convergence of appearance related losses is slowed. Refer to Appendix \ref{Appendix: How pre-training affects loss curves} for more details. Similarily, the 2D object detetcion pre-training, which captures the appearance representation, primarily favors the appearance related losses. These observations confirm this guideline.

Since depth estimation and 2D object detection can imitate the depth and appearance representation of M3OD respectively, a straightforward strategy of conducting M3OD pre-training is combining both them. The experiment in Section \ref{SubSec: Comparing Various Pre-training Tasks} indicates that using each one of them is ineffective, while combining them boosts the M3OD detection performance remarkably. Besides, the reason why we select depth estimation and 2D object detection is that the labels of both them can be obtained without manual effort. Specifically, the depth labels come from the lidar points. The 2D object detection labels can be generated by a well-trained 2D object detector, which is easy to obtain.

In a nutshell, besides the guideline, we develop a basic M3OD pre-training paradigm in this section. In this paradigm, given numerous unlabeled pre-training images, we first project the collected lidar points to the corresponding images for obtaining sparse depth labels, and employ a 2D object detector to generate 2D box labels. Then, we pre-train a model on the pre-training data by conducting depth estimation and 2D object detection simultaneously. After this process, we utilize the backbone parameters of the pre-trained model to initialize the backbone of a M3OD detector and fine-tune this M3OD detector on the M3OD training data.

\vspace{-0.2cm}
\section{DEPT}
\label{Sec: DEPT}
\vspace{-0.2cm}

\begin{figure}[tbp]
    \centering
    \includegraphics[scale=0.32]{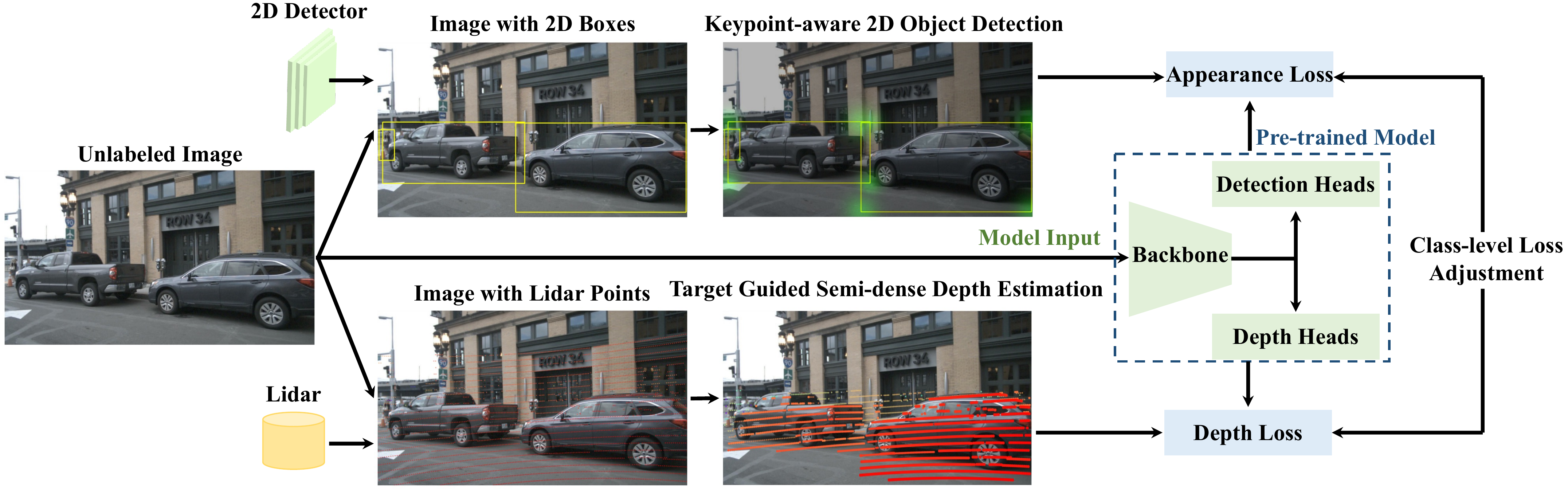}
    \vspace{-0.2cm}
    \caption{The overall pipeline of DEPT.} \label{Fig: Pipeline}
    \vspace{-0.4cm}
\end{figure}

As illustrated in Fig. \ref{Fig: Pipeline}, during the pre-training process, the pre-trained model comprises a backbone and two groups of heads, the depth heads and detection heads. The depth heads are built to learn depth representation and the detection heads are for learning appearance representation. For an unlabeled input image, the sparse depth points are generated by a lidar and the 2D bounding box labels are from a 2D object detector. Therefore, no manual annotation is required. Then, the sparse depth points are transformed as a target guided semi-dense depth map to conduct depth estimation pre-training for the depth heads, which is described in Section \ref{SubSec: Target Guided Semi-dense Depth Estimation}. Meanwhile, the 2D bounding boxes are used to produce keypoint heatmaps, and the heatmaps are combined with the 2D bounding boxes to realize keypoint-aware 2D object detection pre-training. This pre-training process is elaborated in Section \ref{SubSec: Keypoint-aware 2D Object Detection}. Furthermore, as introduced in Section \ref{SubSec: Class-level Loss Adjustment}, to alleviate the class imbalance problem, a class-level loss adjustment strategy is developed to reweight the loss values of various targets.

\vspace{-0.1cm}
\subsection{Target Guided Semi-dense Depth Estimation}
\label{SubSec: Target Guided Semi-dense Depth Estimation}
\vspace{-0.1cm}

In this part, we discuss how we improve the direct depth estimation \cite{liu2020smoke} to imitate the M3OD depth representation. First of all, an M3OD detector only concerns the depth in regions containing targets. Therefore, instead of predicting the depth of lidar points scattered on the whole input image, DEPT trains a model to only estimate the depth in 2D bounding boxes. Secondly, in practical applications, we usually do not care the objects very far from us, and estimating the depth of far targets based on only a monocular image is quite difficult. Hence, the pre-trained model only learns to regress the depth of targets less than 60 meters \cite{ma2021delving}. We name these two operations as region filtering.

\begin{figure}[htbp]
    \centering
    \includegraphics[scale=0.32]{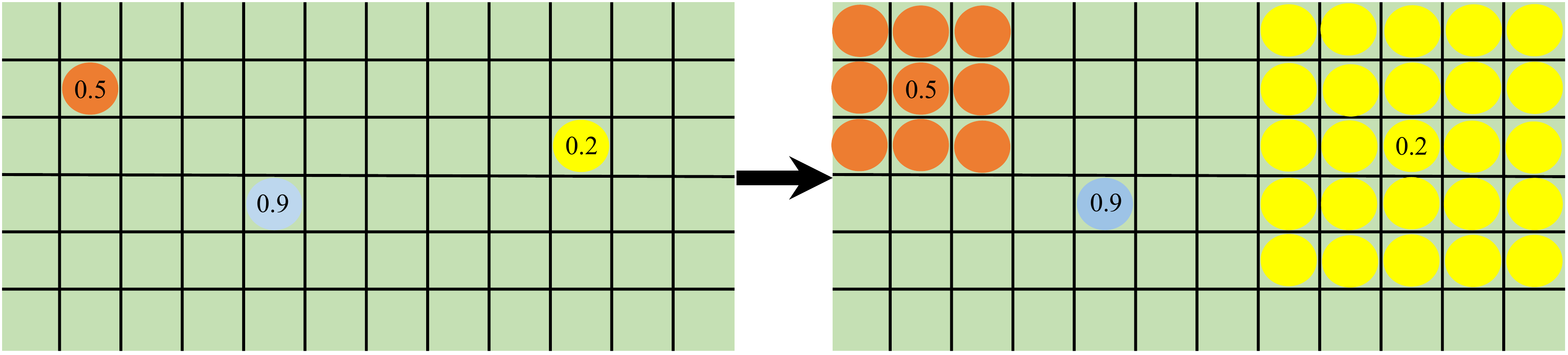}
    \vspace{-0.2cm}
    \caption{The diagram of how a sparse depth map is augmented into a semi-dense depth map. The circles in different colors represent various points on feature maps with depth labels. The number in a circle is the corresponding uncertainty value.} \label{Fig: depth_propagate}
    \vspace{-0.4cm}
\end{figure}

Another issue can be improved is incorporating more pixels for pre-training. Specifically, due to the sparse lidar points, only a small fraction of sparse pixels with depth labels can be exploited for depth estimation pre-training. A possible strategy to alleviate this problem is assigning the depth of the sparse pixels to their surrounding pixels if these sparse pixels are on smooth surfaces of targets. However, if the pixels are near the contours of targets, the depth might change sharply. For these pixels, assigning their depth to the neighboring pixels is inappropriate. Meanwhile, since segmentation annotations are expensive, we do not want to introduce a segmentation model to obtain the masks of targets. Thus, it is difficult to determine whether a pixel is near the contours of targets. To address this obstacle, we model the Laplace uncertainty \cite{kendall2017uncertainties,lu2021geometry} of estimated depth through minimizing the depth loss $L_{dep}$ during the training process:
\begin{align} 
L_{dep} = \frac{\sqrt{2}}{\sigma_{d}} \mid z - z^{gt} \mid + \log \sigma_{d},  \label{Eq1}
\end{align}
where $z$, $z^{gt}$, and $\sigma_{d}$ denote the predicted depth, depth label, and uncertainty value, respectively. In the feature maps extracted by the backbone, since we observe the uncertainty values of points on smooth surfaces are generally smaller, we can use $\sigma_{d}$ to determine the depth of which points are delivered to their neighboring points. In our specific implementation, as shown in Fig. \ref{Fig: depth_propagate},  if the $\sigma_{d}$ of a point is less than 0.3, we propagate its depth to the surrounding 24 points (a 5$\times$5 patch). If $\sigma_{d}$ falls between 0.3 and 0.7, its corresponding depth is assigned to the neighboring 8 points (a 3$\times$3 patch). In this way, we convert a sparse depth map into a semi-dense depth map.

\vspace{-0.1cm}
\subsection{Keypoint-aware 2D Object Detection}
\label{SubSec: Keypoint-aware 2D Object Detection}
\vspace{-0.1cm}

For a M3OD model, the ability of recognizing keypoints of targets is important because the positions of keypoints contain the information of dimension and orientation. In addition, some M3OD detectors directly utilize keypoints to calculate the depth of targets \cite{li2021monocular}. Therefore, besides the basic 2D object detection, we hope to add a subtask to boost the sensitivity of the pre-trained model to keypoints. 

To this end, we produce keypoint response heatmaps of 4 channels based on the 4 corner keypoints of a 2D bounding box, which is illustrated in Fig. \ref{Fig: Pipeline}. Every corner keypoint corresponds to a response heatmap channel. The keypoint response heatmaps are generated in a way similar to CenterNet \cite{zhou2019objects}, where the representative point of each target is modeled with a Gaussian distribution:
\begin{align}
Y_{xyc} = \exp(-\frac{(x_{c}-p_{c}^{x})^{2} + (y_{c}-p_{c}^{y})^{2}}{2\sigma^{2}_{p}}), \label{Eq2}
\end{align}
where $Y_{xyc}$ is the heatmap value at the $c$th channel with the coordinate $(x, y)$. $(p_{c}^{x}, p_{c}^{y})$ and $\sigma_{p}$ represent the keypoint coordinate at the $c$th channel and an object size-adaptive standard deviation. Notably, we omit the downsampling process by the backbone in Eq. \ref{Eq2} for simplicity. We build a special head on the pre-trained backbone to regress the keypoint heatmaps, and the network is updated using Focal loss \cite{lin2017focal}. In this way, the sensibility of the pre-trained model to keypoints is improved.

\vspace{-0.1cm}
\subsection{Class-level Loss Adjustment}
\label{SubSec: Class-level Loss Adjustment}
\vspace{-0.1cm}

The data in autonomous driving datasets is highly imbalanced. For example, there exist 513462 cars while only 11154 bicycles in the training set of nuScenes. According to the information theory \cite{reza1994introduction}, the knowledge contained in an object belonging to rare classes is more valuable. Thus, we reweight the loss of every class based on a class-level loss adjustment strategy, which is shown in Fig. \ref{Fig: Pipeline}.

We compute the weight of each class by counting its sample number in the training set. Specifically, assume there are $n$ classes and the sample number of the $k{\rm th}$ class is $s_{k}$. Let 
\begin{align}
s_{m} = \max(s_{1}, s_{2}, ..., s_{n}), \label{Eq3}
\end{align}
the weight for the $k{\rm th}$ class can be defined as:
\begin{align}
w_{k} = \sqrt{\frac{s_{m}}{s_{k}}}. \label{Eq4}
\end{align}
Through reweighting the loss of the $k{\rm th}$ class with $w_{k}$, a more balanced representation is obtained.

\vspace{-0.2cm}
\section{Experiments}
\label{Sec: Experiments}
\vspace{-0.2cm}

In this section, we conduct experiments on KITTI and nuScenes to evaluate DEPT. The process of obtaining an M3OD detector comprises two phases, pre-training and fine-tuning. For the first phase, the model is pre-trained for 6 epochs with the learning rate (LR) of 1e-4. In the second phase, the model is fine-tuned for 14 epochs on nuScenes (LR is 1e-4) or 100 epochs on KITTI-3D (LR is 2.25e-4). All the experiments are conducted on a server with 4 RTX3090 GPUs. In the following, we introduce the two employed benchmarks and the baseline detector.

\vspace{-0.1cm}
\paragraph{KITTI.} The KITTI 3D object detection benchmark (KITTI-3D) \cite{geiger2012we} consists of 7481 images for training and 7518 images for testing. Following \cite{zhou2018voxelnet}, the training images are further divided into the training set (3712 images) and validation set (3769 images). In KITTI, there are mainly three classes of targets needing to be considered, Car, Pedestrian, and Cyclist. These targets are categorized into three difficulty levels (Easy, Moderate, and Hard) according to their pixel heights, etc. The average precision (AP) based on 3D Intersection over Union (IOU) is the primary evaluation metric. For the collecting data platform, each vehicle is equipped with 2 front-facing color cameras and 1 lidar. Although only the 3D bounding box labels of the 7481 images in KITTI-3D are available, there are another 40282 unlabeled images with lidar points, known as KITTI-raw. Following DD3D \cite{park2021pseudo}, we remove the images in KITTI-raw that are geographically close to any image in KITTI-3D. The remaining images are used for pre-training.

\vspace{-0.1cm}
\paragraph{nuScenes.} Compared with KITTI, the nuScenes benchmark \cite{caesar2020nuscenes} contains more challenging data. It includes 1000 scenes and every scene is roughly 20 seconds in length. During the collecting data process, every vehicle is equipped with 1 lidar and 6 cameras facing in different directions. The images are captured at 12Hz and annotated at 2Hz. Thus, there are numerous unlabeled data between annotated frames, namely sweep images. Nevertheless, the semantic information of these sweep images is highly similar to the annotated ones and contributes little to the pre-training process. In nuScenes, 10 categories of objects are considered. The evaluation metrics include mean average precision (mAP) based on center distance, true positive metrics (TP metrics), and nuScenes detection score (NDS). The TP metrics can be further divided into ATE, ASE, AOE, AVE, and AAE, which are for reflecting the performance of estimating translation, dimension, orientation, velocity, and attribute, respectively. The NDS is computed based on the mAP and TP metrics.

\vspace{-0.1cm}
\paragraph{Baseline Detector.}
We build the baseline detector based on MonoFlex \cite{zhang2021objects}, which comprises a backbone (e.g., DLA34 \cite{yu2018deep}) and several heads. Every head consists of two convolutional layers and a batch normalization layer. MonoFlex predicts the depth of targets by combining direct depth estimation and calculating depth using height. Although this fusion strategy is effective for boosting the detection scores on KITTI, it degrades the performance on nuScenes. Through analysis, we find this is because the scenes in nuScenes are more challenging and the height of targets is difficult to estimate. Since nuScenes reflects practical challenges better, we remove the calculating depth using height part from the baseline. This modification decreases the performance on KITTI to some extent. Besides, when evaluated on nuScenes, two heads are added for estimating the attribute and speed of targets following the implementation in DD3D \cite{park2021pseudo}.

\vspace{-0.1cm}
\subsection{Benchmark Evaluation}
\label{SubSec: Benchmark Evaluation}
\vspace{-0.1cm}

In this part, we evaluate our method on two benchmarks, KITTI-3D and nuScenes. For KITTI-3D, we first pre-train a backbone (DLA34) with DEPT using KITTI-raw. The performances of the baseline detectors loading and without loading the pre-trained backbone are presented in Table \ref{Table: Performance comparison on the KITTI-3D testing set} and Table \ref{Table: Performance comparison on the KITTI-3D validation set}, which correspond to the evaluation results on the KITTI-3D testing and validation sets, respectively.

\begin{table}[htbp]
\vspace{-0.2cm}
  \caption{Analyzing the effect of loading the pre-trained backbone on the KITTI-3D testing set.}
  \label{Table: Performance comparison on the KITTI-3D testing set}
  \centering
  \resizebox{125mm}{10mm}{
  \begin{tabular}{cccccccccc}
    \toprule
    \multirow{2}{*}{Pre-trained} & \multicolumn{3}{c}{Car, ${\rm AP}_{3D}70$ (\%) $\uparrow$} & \multicolumn{3}{c}{Pedestrian, ${\rm AP}_{3D}70$ (\%) $\uparrow$} & \multicolumn{3}{c}{Cyclist, ${\rm AP}_{3D}70$ (\%) $\uparrow$} \\
    \cmidrule(r){2-10}
    & Easy & Moderate & Hard & Easy & Moderate & Hard & Easy & Moderate & Hard \\
    \cmidrule(r){1-10}
    No & 19.15 & 11.65 & 9.92 & 13.08 & 8.35 & 6.88 & 4.34 & 2.31 & 2.38 \\
    Yes & 20.43 & 13.83 & 11.66 & 16.28 & 10.81 & 9.15 & 7.67 & 4.29 & 3.33 \\
    \bottomrule
  \end{tabular}}
\vspace{-0.2cm}
\end{table}

\begin{table}[htbp]
\vspace{-0.2cm}
  \caption{Analyzing the effect of loading the pre-trained backbone on the KITTI-3D validation set.}
  \label{Table: Performance comparison on the KITTI-3D validation set}
  \centering
  \resizebox{125mm}{10mm}{
  \begin{tabular}{cccccccccc}
    \toprule
    \multirow{2}{*}{Pre-trained} & \multicolumn{3}{c}{Car, ${\rm AP}_{3D}70$ (\%) $\uparrow$} & \multicolumn{3}{c}{Pedestrian, ${\rm AP}_{3D}70$ (\%) $\uparrow$} & \multicolumn{3}{c}{Cyclist, ${\rm AP}_{3D}70$ (\%) $\uparrow$} \\
    \cmidrule(r){2-10}
    & Easy & Moderate & Hard & Easy & Moderate & Hard & Easy & Moderate & Hard \\
    \cmidrule(r){1-10}
    No & 19.17 & 14.46 & 12.72 & 8.48 & 6.59 & 5.50 & 5.72 & 2.99 & 2.74 \\
    Yes & 22.43 & 17.28 & 14.71 & 8.89 & 6.61 & 5.23 & 7.76 & 4.16 & 3.83 \\
    \bottomrule
  \end{tabular}}
\vspace{-0.2cm}
\end{table}

According to the results in Table \ref{Table: Performance comparison on the KITTI-3D testing set} and Table \ref{Table: Performance comparison on the KITTI-3D validation set}, the performance of the baseline detector is improved significantly by loading the backbone pre-trained on KITTI-raw. For example, as presented in Table \ref{Table: Performance comparison on the KITTI-3D testing set}, the ${\rm AP}_{3D}70$ score of the moderate level is boosted by 18.71\% relatively (from 11.65 to 13.83).

Furthermore, we verify the effectiveness of DEPT on the nuScenes validation set. We do not conduct this experiment on the nuScenes testing set because only 3 submissions are permitted per year. Since the effectiveness of pre-training relies on incorporating extra unlabeled data heavily, we employ the whole training set of nuScenes for pre-training without using 3D box labels. Afterwards, the pre-trained model is fine-tuned on 6000 samples from the training set with 3D box labels and then tested on the validation set. For accelerating the training process, the resolution of input images is downsampled as $(1024, 576)$. The experimental results are reported in Table \ref{Table: Performance comparison on the nuScenes validation set}. The results on the metrics mAVE and AOE are omitted because we mainly concern the quality of the estimated 3D boxes.

\begin{table}[htbp]
\vspace{-0.2cm}
  \caption{Analyzing the effect of loading a pre-trained backbone on the nuScenes validation set.}
  \label{Table: Performance comparison on the nuScenes validation set}
  \centering
  \resizebox{100mm}{9mm}{
  \begin{tabular}{cc|ccccc}
    \toprule
    Pre-trained & Whole data & mAP $\uparrow$ & mATE $\downarrow$ & mASE $\downarrow$ & mAOE $\downarrow$ & NDS $\uparrow$ \\
    \hline
    & & 0.1449 & 0.8089 & 0.4113 & 0.7206 & 0.2346 \\
    \hline
    \checkmark & & 0.1566 & 0.7738 & 0.4153 & 0.6922 & 0.2476 \\
    \hline
    \checkmark& \checkmark & 0.2207 & 0.7681 & 0.2810 & 0.7124 & 0.3294 \\
    \bottomrule
  \end{tabular}}
\vspace{-0.2cm}
\end{table}

The first row of results in Table \ref{Table: Performance comparison on the nuScenes validation set} corresponds to the model without loading the pre-trained backbone. and the second row represents the model loading the backbone pre-trained on the 6000 samples used in the fine-tuning phase, which means the pre-training images are the same as the fine-tuning images. Comparing these two rows, we can observe the pre-trained backbone improves the performance slightly. By contrast, as shown in the third row, when we use the whole nuScenes training set (28130 samples) to pre-train the backbone, the performance of the model is boosted by a large margin. For instance, the NDS score is improved by 40.4\% relatively (from 0.2346 to 0.3294). Therefore, the used data volume is critical for pre-training. Besides, as suggested by the results in Table \ref{Table: Comparison between different pre-training tasks}, DEPT is more efficient than DD3D \cite{park2021pseudo}, which completely relies on direct depth estimation. Hence, the key value of DEPT is \textbf{it provides a more efficient way than DD3D to utilize the knowledge of more data without extra manual annotation.}

\vspace{-0.1cm}
\subsection{Comparing Various Pre-training Tasks}
\label{SubSec: Comparing Various Pre-training Tasks}
\vspace{-0.1cm}

In this experiment, we compare how various pre-training tasks affect the M3OD performance on KITTI. The studied pre-training tasks include CL, MAE \cite{he2021masked}, depth estimation, and 2D object detection. Among them, CL is implemented based on MoCo \cite{he2020momentum}. Although MAE is designed for transformer \cite{carion2020end}, pre-training CNN with MAE is meaningful since CNN can also restore masked images, as illustrated in Appendix \ref{Appendix: MAE Reconstruction results of CNN}. In depth estimation, we project the lidar points to the corresponding 2D camera images, and train a network to predict the depth of pixels with projected lidar points. For 2D object detection, we first train a 2D detector (e.g., CenterNet \cite{zhou2019objects}) with the training set of KITTI-3D, and then employ the trained detector to generate 2D boxes for all unlabeled images in KITTI-raw. 

We regard KITTI-raw as the pre-training data. When the pre-training phase is completed, only the backbone of the pre-trained network is retained for initializing the backbone weights of the monocular 3D object detector. Then, the monocular 3D object detector is fine-tuned on the training set of KITTI-3D and evaluated on the validation set. The experimental results are presented in Table \ref{Table: Comparison between different pre-training tasks}. `Depth' and `Det' refer to the depth estimation and 2D object detection pre-training tasks.

\begin{table}[htbp]
\vspace{-0.2cm}
  \caption{Comparison between different pre-training tasks.}
  \label{Table: Comparison between different pre-training tasks}
  \centering
  \resizebox{135mm}{18mm}{
  \begin{tabular}{lccccccccc}
    \toprule
    \multirow{2}{*}{Pre-training} & \multicolumn{3}{c}{Car, ${\rm AP}_{3D}70$ (\%) $\uparrow$} & \multicolumn{3}{c}{Pedestrian, ${\rm AP}_{3D}70$ (\%) $\uparrow$} & \multicolumn{3}{c}{Cyclist, ${\rm AP}_{3D}70$ (\%) $\uparrow$} \\
    \cmidrule(r){2-10}
    & Easy & Moderate & Hard & Easy & Moderate & Hard & Easy & Moderate & Hard \\
    \cmidrule(r){1-10}
    None & 19.17 & 14.46 & 12.72 & 8.48 & 6.59 & 5.50 & 5.72 & 2.99 & 2.74 \\
    CL & 17.05 & 12.12 & 10.55 & 7.52 & 6.02 & 4.83 & 2.90 & 1.66 & 1.52 \\
    MAE & 9.68 & 6.73 & 5.52 & 5.57 & 3.76 & 3.27 & 0.42 & 0.22 & 0.23 \\
    Depth & 18.15 & 13.70 & 12.19 & 11.47 & 8.32 & 6.83 & 4.49 & 2.67 & 2.14 \\
    Det & 20.45 & 14.88 & 13.26 & 10.43 & 8.16 & 6.58 & 9.37 & 5.15 & 4.54 \\
    Depth+Det & 22.02 & 16.30 & 13.96 & 12.25 & 8.96 & 6.97 & 10.26 & 5.68 & 5.45 \\
    \bottomrule
  \end{tabular}}
\vspace{-0.2cm}
\end{table}

As shown in Table \ref{Table: Comparison between different pre-training tasks}, compared with the model without pre-training, both CL and MAE degrade the performance, and MAE decreases the APs dramatically, which implies that the learned representation is unsuitable for M3OD. According to the 1st, 4th, and 5th rows of results in Table \ref{Table: Comparison between different pre-training tasks}, depth estimation and 2D object detection do not improve the performance significantly. This observation is inconsistent with the one in DD3D \cite{park2021pseudo}. We speculate this is because DD3D incorporates numerous external data (about 15M frames) for pre-training, which boosts the detection precision effectively. Through analyzing the loss curves, we observe that the depth estimation pre-training accelerates the convergence of depth related losses significantly but slows the unrelated ones. Conversely, the 2D object detection pre-training favors the convergence of appearance related losses, such as the orientation loss. Hence, we speculate that a pre-training task benefits the learning of downstream tasks similar to it while harming the dissimilar ones. Meanwhile, comparing the 1st and 6th rows of results in Table \ref{Table: Comparison between different pre-training tasks}, we can note that combining depth estimation and 2D object detection is a very efficient pre-training paradigm.

\vspace{-0.1cm}
\subsection{Ablation Study}
\label{SubSec: Ablation Study}
\vspace{-0.1cm}

In this part, we validate the effectiveness of various strategies described in Section \ref{Sec: DEPT} on nuScenes. We do not conduct this study on KITTI because we observe some performance randomness from the results of KITTI, which may be because the data volume of KITTI is not very large. Thus, if the improvement of a strategy is not quite significant, its effectiveness is difficult to confirm on KITTI. 

Following Section \ref{SubSec: Benchmark Evaluation}, this experiment employs the whole nuScenes training set to pre-train a backbone and uses 6000 samples in the training set to fine-tune models loading this backbone. The studied strategies include region filtering (RF) in Section \ref{SubSec: Target Guided Semi-dense Depth Estimation}, keypoint heatmap estimation (KHE) in Section \ref{SubSec: Keypoint-aware 2D Object Detection}, class-level loss adjustment (CLA) in Section \ref{SubSec: Class-level Loss Adjustment}, and semi-dense depth estimation (SDE) in Section \ref{SubSec: Target Guided Semi-dense Depth Estimation}. The experimental results are reported in Table \ref{Table: Ablation study on the various strategies}.

\begin{table}[htbp]
\vspace{-0.2cm}
  \caption{Ablation study of the various strategies applied to the baseline paradigm on nuScenes.}
  \label{Table: Ablation study on the various strategies}
  \centering
  \resizebox{125mm}{15mm}{
  \begin{tabular}{ccccc|ccccc}
    \toprule
    Pre-train & RF & KHE & CLA & SDE & mAP $\uparrow$ & mATE $\downarrow$ & mASE $\downarrow$ & mAOE $\downarrow$ & NDS $\uparrow$ \\
    \hline
    & & & & & 0.1449 & 0.8089 & 0.4113 & 0.7206 & 0.2346 \\
    \hline
    \checkmark & & & & & 0.2089 & 0.7831 & 0.2881 & 0.8134 & 0.2922 \\
    \hline
    \checkmark & \checkmark & & & & 0.2104 & 0.7821 & 0.2886 & 0.8149 & 0.2923 \\
    \hline
    \checkmark & \checkmark & \checkmark & & & 0.2125 & 0.7753 & 0.2818 & 0.7332 & 0.3177 \\
    \hline
    \checkmark & \checkmark & \checkmark & \checkmark & & 0.2156 & 0.7722 & 0.2800 & 0.7111 & 0.3234 \\
    \hline
    \checkmark & \checkmark & \checkmark & \checkmark & \checkmark & 0.2207 & 0.7681 & 0.2810 & 0.7124 & 0.3294 \\
    \bottomrule
  \end{tabular}}
\vspace{-0.2cm}
\end{table}

As presented in Table \ref{Table: Ablation study on the various strategies}, all the applied strategies are effective. Specifically, the first row of results in Table \ref{Table: Ablation study on the various strategies} shows the performance of the detector without pre-training, and the second row corresponds to the detector loading the backbone pre-trained using the baseline paradigm. We can observe from these two rows that the baseline paradigm improves the performance on all metrics by large margins except mAOE, which reflects the capability of a model on predicting orientation. This observation reveals that both the depth estimation and 2D object detection fail to imitate the representation of orientation estimation. However, when KHE is applied, the error of estimating orientation (mAOE) is decreased significantly, which suggests that the representation of KHE is similar to orientation estimation. This finding can be explained by the insight that the orientation angle is able to be computed based on the locations of eight 3D corner keypoints.

\vspace{-0.1cm}
\subsection{Alleviating the Long-tailed Distribution}
\label{SubSec: Alleviating the Long-tailed Distribution}
\vspace{-0.1cm}

The long-tailed distribution problem is serious in various practical visual perception applications. For example, in KITTI-3D, there are actually 8 classes of targets, while we usually only consider the first 3 classes, i.e., Car, Pedestrian, and Cyclist. This is mainly because the target numbers of the remaining 5 classes are two few. To reveal the seriousness of the long-tailed distribution problem on autonomous driving, we employ MonoFlex to detect all the 8 classes of targets in KITTI-3D. The detection results of the 8 classes on the KITTI-3D validation set and the target numbers of all classes from the KITTI-3D training set are illustrated in Fig. \ref{Fig: Long-tailed distribution on KITTI}. Two observations can be drawn form Fig. \ref{Fig: Long-tailed distribution on KITTI}. First of all, the detection score of a class decreases as the training sample of this class becomes fewer. Secondly, the performance of some  especially rare classes is quite poor. For instance, the scores of Truck, Person\_sit, and Tram arrive zero. 

\begin{figure}[htbp]
    \centering
    \includegraphics[scale=0.25]{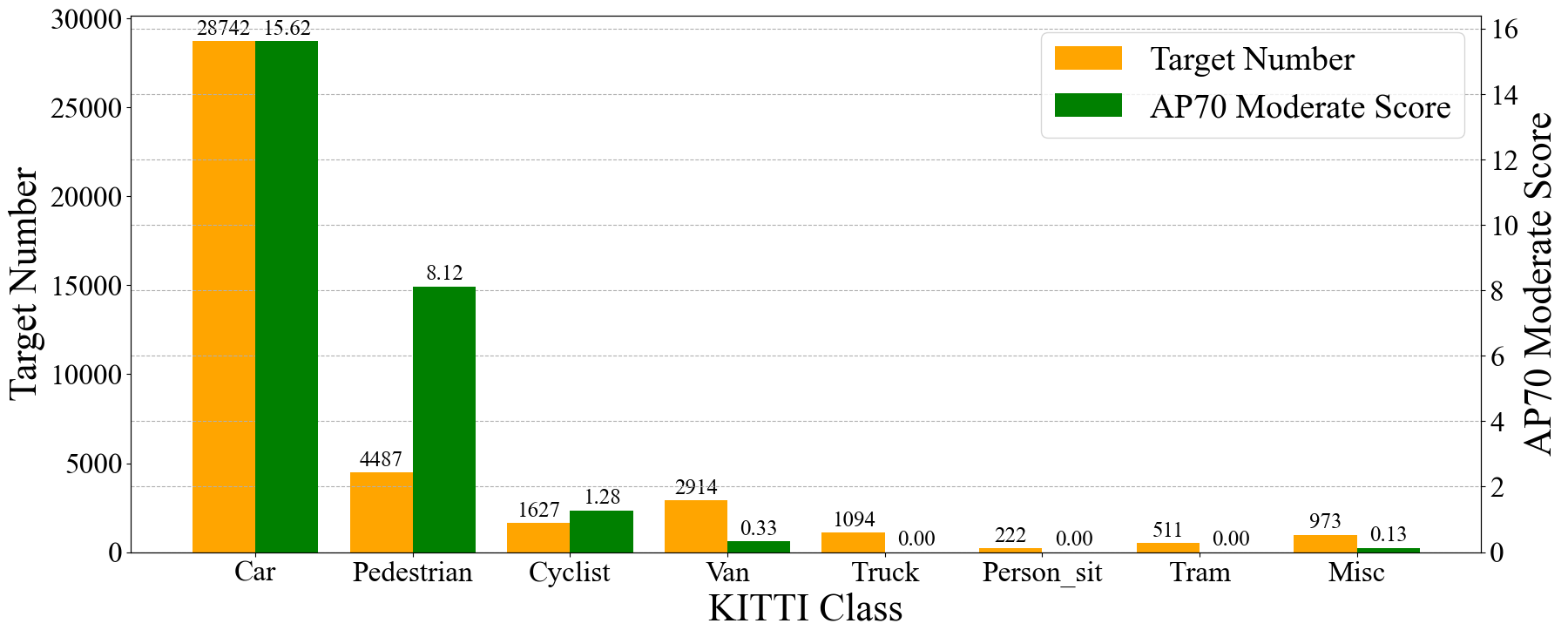}
    \vspace{-0.2cm}
    \caption{The target numbers of the 8 classes in the KITTI-3D training set and the corresponding detection results of MonoFlex on these 8 classes evaluated on the KITTI-3D validation set.} \label{Fig: Long-tailed distribution on KITTI}
    \vspace{-0.2cm}
\end{figure}

Interestingly, we note that DEPT can alleviate the long-tailed distribution problem effectively. For example, as reported in Table \ref{Table: Performance comparison on the KITTI-3D testing set}, considering the Moderate level, the ${\rm AP}_{3D}70$ scores of Car, Pedestrian, and Cyclist are improved by 18.7\%, 29.5\%, and 85.71\% relatively, respectively. Thus, DEPT benefits the classes of rare samples more than the classes with abundant samples. To further confirm this observation, we also analyze the evaluation results of various classes on the nuScenes validation set, which is given in Table \ref{Table: mAP scores of various classes on the nuScenes validation set}. Since the object numbers of these 10 classes in the training set are 513642, 91122, 15984, 27560, 15775, 213207, 11763, 11154, 91770, 149656, respectively, we can still find that the classes of fewer targets are improved by DEPT with larger relative margins.

\begin{table}[htbp]
  \caption{The mAP scores of various classes on the nuScenes validation set.}
  \label{Table: mAP scores of various classes on the nuScenes validation set}
  \centering
  \resizebox{140mm}{7.5mm}{
  \begin{tabular}{ccccccccccc}
    \toprule
    Pre-trained & Car & Truck & Bus & Trailer & Constru. & Pedestrain & Motorcycle & Bicycle & Traffi. & Barrier \\
    \cmidrule(r){1-11}
    No & 0.306 & 0.043 & 0.044 & 0.001 & 0.001 & 0.276 & 0.130 & 0.100 & 0.353 & 0.196 \\
    Yes & 0.391 & 0.107 & 0.167 & 0.282 & 0.015 & 0.351 & 0.216 & 0.178 & 0.435 & 0.319 \\
    \bottomrule
  \end{tabular}}
\vspace{-0.2cm}
\end{table}

\vspace{-0.1cm}
\subsection{Generalization between Domains}
\label{SubSec: Generalization between Domains}
\vspace{-0.1cm}

The generalization ability of the obtained backbone is critical for pre-training, because the backbone is expected be applied to various domains. In this experiment, we aim to verify this issue through first pre-training a model on the nuScenes training set and then fine-tuning and validating it on KITTI-3D. The results are reported in Table \ref{Table: Generalization ability analysis}.

\begin{table}[htbp]
\vspace{-0.2cm}
  \caption{Analyzing the generalization ability of the pre-trained representation on KITTI-3D.}
  \label{Table: Generalization ability analysis}
  \centering
  \resizebox{130mm}{10mm}{
  \begin{tabular}{cccccccccc}
    \toprule
    \multirow{2}{*}{Pre-trained on nuScenes} & \multicolumn{3}{c}{Car, ${\rm AP}_{3D}70$ (\%) $\uparrow$} & \multicolumn{3}{c}{Pedestrian, ${\rm AP}_{3D}70$ (\%) $\uparrow$} & \multicolumn{3}{c}{Cyclist, ${\rm AP}_{3D}70$ (\%) $\uparrow$} \\
    \cmidrule(r){2-10}
    & Easy & Moderate & Hard & Easy & Moderate & Hard & Easy & Moderate & Hard \\
    \cmidrule(r){1-10}
    No & 19.17 & 14.46 & 12.72 & 8.48 & 6.59 & 5.50 & 5.72 & 2.99 & 2.74 \\
    Yes & 21.39 & 16.51 & 13.90 & 12.65 & 9.61 & 7.28 & 6.36 & 3.94 & 3.38 \\
    \bottomrule
  \end{tabular}}
\vspace{-0.2cm}
\end{table}

Comparing the 1st and 2nd rows of results in Table \ref{Table: Generalization ability analysis}, we can observe that the backbone pre-trained on nuScenes still benefits the performance on KITTI-3D, although there exists a significant domain gap between these two benchmarks. Thus, the representation obtained from DEPT generalizes well.

\vspace{-0.2cm}
\section{Conclusion and Limitation}
\label{Sec: Conclusion and Limitation}
\vspace{-0.2cm}

In this work, we first have explored the guideline of devising pre-training tasks. Afterwards, following the obtained guideline, an efficient M3OD pre-training paradigm without requiring extra manual annotation has been proposed. Through extensive experiments on KITTI and nuScenes, we have confirmed that this paradigm can improve the performance and generalization ability of a M3OD detector significantly. Besides, it is also effective for addressing the M3OD long-tailed distribution problem. We hope this work can shed light on how to utilize the great amount of unlabeled data in autonomous driving applications. The main limitation of this work is we do not try pre-training large models because we do not have enough pre-training data. We believe a more powerful model with a better generalization ability would be obtained if DEPT is deployed in practical autonomous driving applications, where exists numerous unlabeled data.

{
\bibliographystyle{plain} 
\bibliography{reference}
}

\appendix

\section{Appendix}
\label{Appendix}

In this section, we put more discussion and experimental results about this work.

\subsection{Comparison with DD3D}
\label{Appendix: Comparison with DD3D}

DEPT differs from DD3D in the used pre-training tasks. In the manuscript, we do not directly compare the backbones obtained by DEPT and DD3D, because DD3D uses numerous extra private data. However, we compare the effectiveness of their pre-training strategies in Table \ref{Table: Comparison between different pre-training tasks}, where ``Depth'' refers to DD3D and ``Depth+Det'' is the baseline form of DEPT. According to the results in Table \ref{Table: Comparison between different pre-training tasks}, the pre-training strategy of DEPT is much more efficient than the one of DD3D.

\subsection{Performance of the Baseline}
\label{Appendix: Performance of the Baseline}

Our baseline detector is implemented based on the source code of MonoFlex. As shown in Table \ref{Table: Performance comparison on the KITTI-3D testing set}, the scores of the baseline detector on the KITTI-3D testing set are lower than the results reported by MonoFlex. This phenomenon is mainly attributed to two reasons: (1) As stated in Section \ref{Sec: Experiments}, we remove the calculating depth using height part from MonoFlex because we observe this strategy degrades the performance on nuScenes, and we hope to keep a unified baseline detector between KITTI and nuScenes. (2) Even though we do not apply any modification to the source code of MonoFlex, we still cannot reproduce its reported performance on the testing set. The reason is unclear. We speculate that this phenomenon may be caused by some issues like we use a different type of GPU, because some GPUs have better computing precision and stability. 

Nevertheless, since we only want to confirm the effectiveness of our method, this problem does not matter if we keep the same baseline.

\subsection{How Depth Pre-training Affects Loss Curves}
\label{Appendix: How pre-training affects loss curves}

\begin{figure}[htbp]
    \centering
    \includegraphics[scale=0.36]{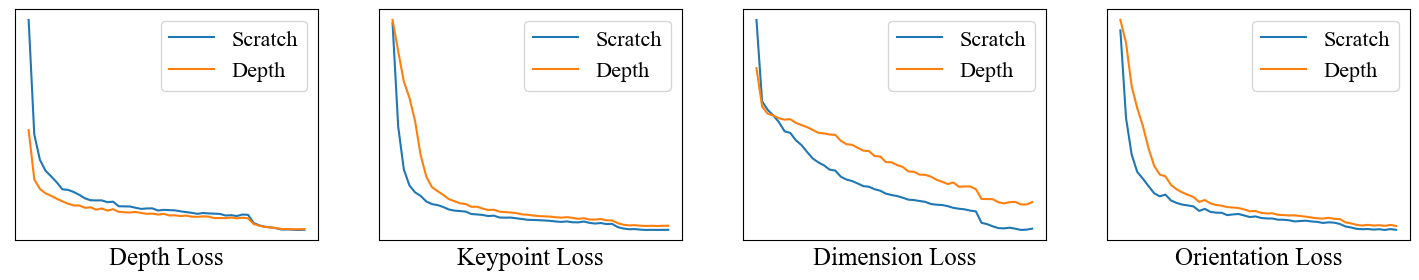}
    \caption{The loss curves of models with and without depth estimation pre-training.} \label{Fig: DEPT_depth_loss}
    \vspace{-0.4cm}
\end{figure}

In this part, we analyze how depth pre-training influences the convergence of various loss items of M3OD training. Specifically, we compare the loss curves with and without depth estimation pre-training. For the model with pre-training, we first pre-train it by depth estimation on the training set of KITTI-3D using sparse depth labels. Afterwards, we fine-tune this model on the KITTI-3D training set to learn to detect 3D targets. Conversely, for the model without pre-training, it is directly trained on the KITTI-3D training set to produce 3D boxes containing targets. We select four loss items (depth loss, keypoint loss, dimension loss, and orientation loss) and illustrate them in Fig. \ref{Fig: DEPT_depth_loss}.

As shown, the depth pre-training benefits the convergence of depth loss, while harming the convergence of the remaining losses, which are mainly related to the appearance representation. We have also observed a similar phenomenon from the 2D object detection pre-training, which primarily favors the convergence of losses related to appearance representation. The visualization results further confirm our guideline of devising pre-training tasks.

Besides, there is another notable phenomenon. As shown in Fig. \ref{Fig: DEPT_depth_loss}, although the depth pre-training accelerates the convergence of the depth loss at the first several epochs, the loss values of models with and without depth estimation pre-training gradually converge to the same value. Similar phenomenons can also be observed from the loss curves of keypoint and orientation. Through analysis, we find this is because the pre-training data is the same as the fine-tuning data. In other words, no extra information from new data is used. These observations indicate that incorporating numerous extra data is critical for pre-training, because the essence of pre-training is like a tool to exploit information in new data with little extra manual effort.

\subsection{MAE Reconstruction Results of CNN}
\label{Appendix: MAE Reconstruction results of CNN}

In Section \ref{SubSec: Comparing Various Pre-training Tasks}, we pre-train a model on KITTI-raw using MAE. Nevertheless, the experimental results indicate that this pre-training process degrades the performance on the KITTI-3D validation set. This phenomenon could be caused by two reasons: (1) The representation learned by MAE is dissimilar to the representation of M3OD. (2) Since MAE is a task designed for pre-training Transformer, CNN fails to learn to recover masked images through MAE. 

\begin{figure}[htbp]
    \centering
    \includegraphics[scale=0.48]{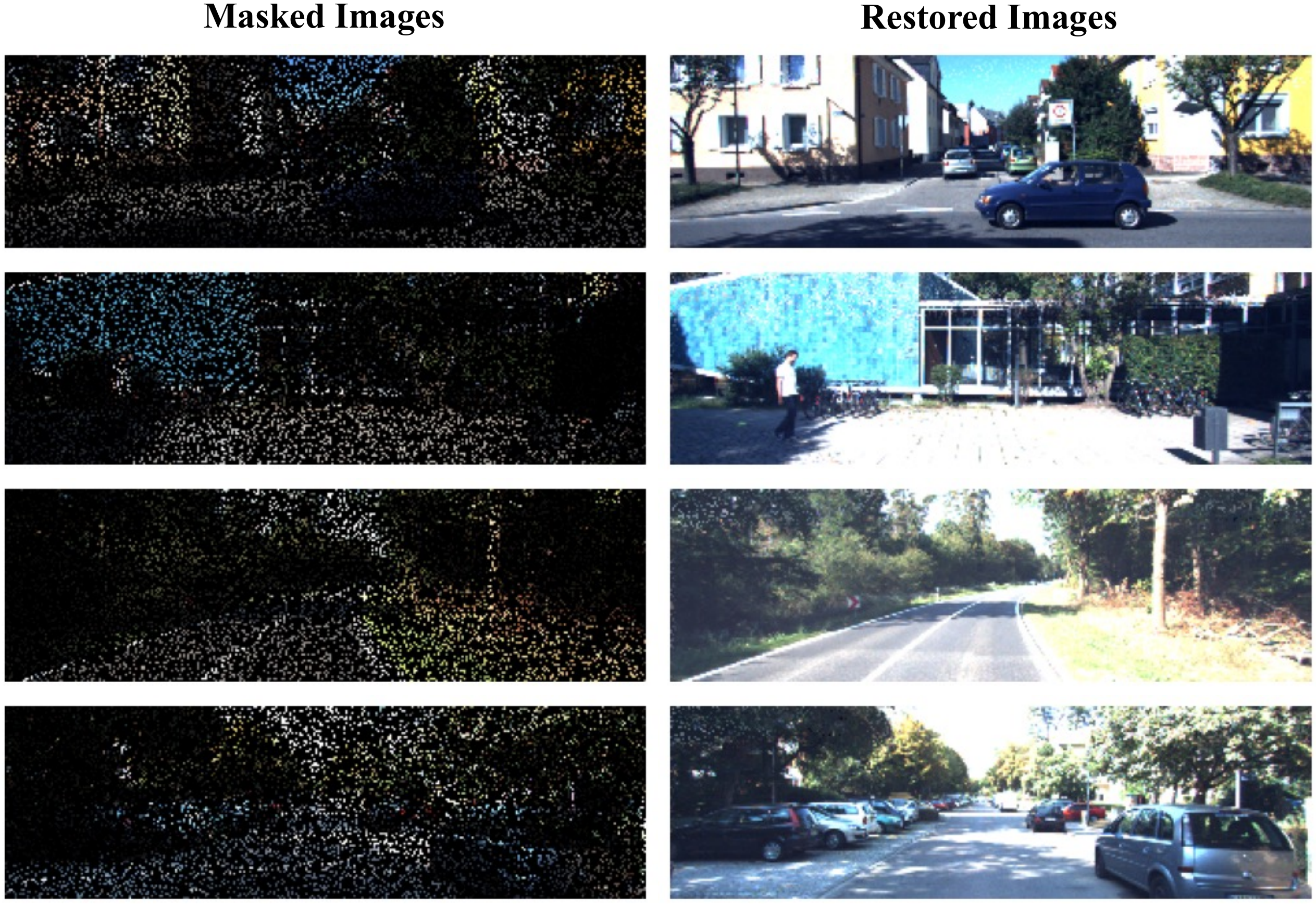}
    \caption{Visualization of MAE. The left column is the randomly masked images and the right column corresponds to the images restored by a pre-trained CNN model.} \label{Fig: Long-tailed distribution on KITTI} \label{Fig: MAE visualization}
    \vspace{-0.4cm}
\end{figure}

To confirm which one is the real cause, we randomly mask 80\% pixels of images from the KITTI-3D validation set, and use the model pre-trained on KITTI-raw with the MAE pre-training task to recover them. Notably, all the images in KITTI-raw that are geographically close to any image in KITTI-3D have been removed. Several masked input images and the restored images are visualized in Fig. \ref{Fig: MAE visualization}. As shown, CNN recover the masked images successfully. Therefore, we think the unsatisfactory performance of the fine-tuned model is because MAE is unsuitable as an M3DO pre-training task.

\subsection{Visualization of Detection Results}
\label{Appendix: Visualization of Detection Results}

\begin{figure}[htbp]
\vspace{-0.2cm}
    \centering
    \includegraphics[scale=0.47]{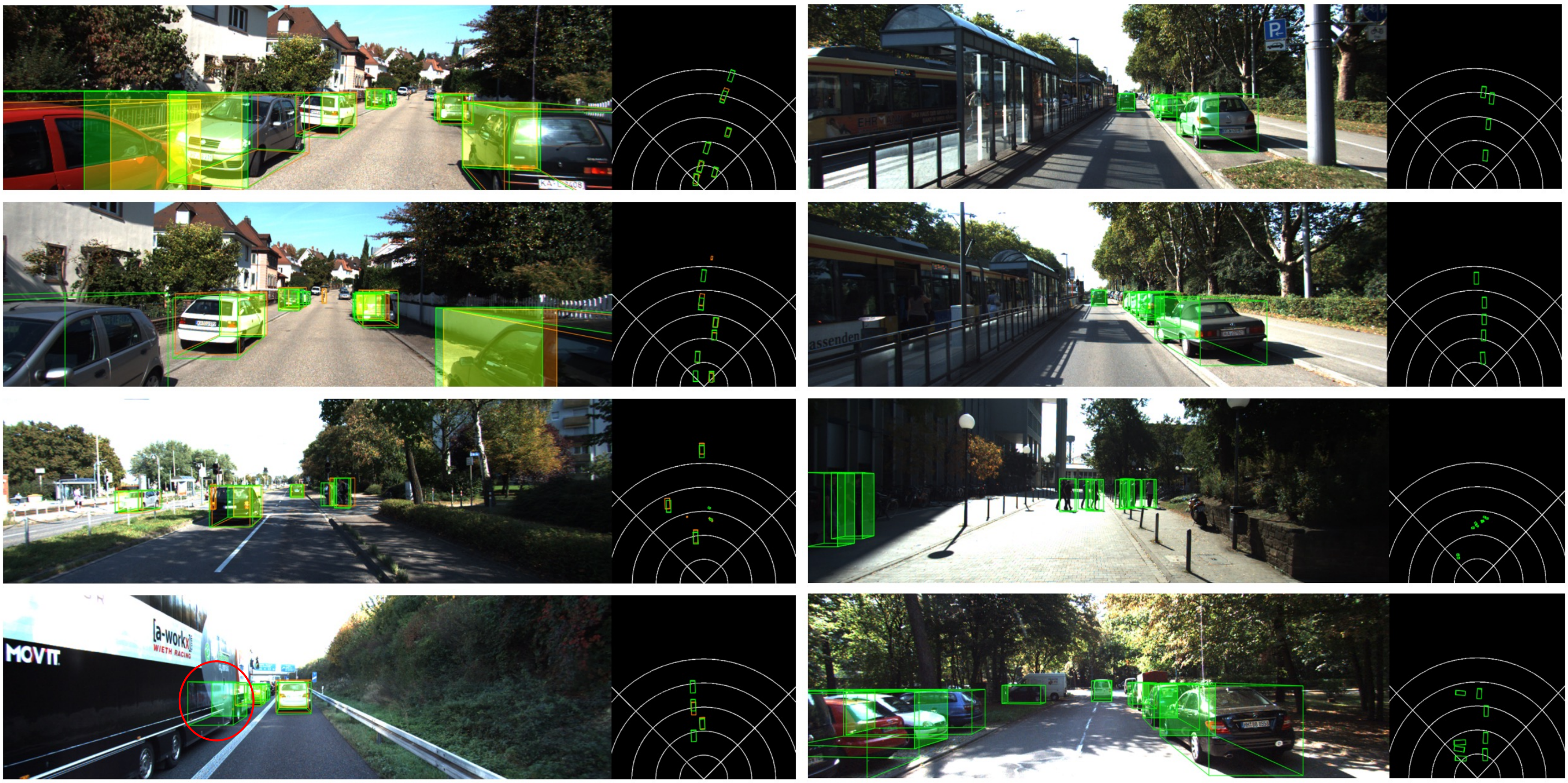}
    \vspace{-0.2cm}
    \caption{Visualization of some detection results on KITTI-3D. The left column is the validation set and the right column corresponds to the testing set. The estimation results of the model are in green and the ground truth boxes are marked in orange.} \label{Fig: Detection results on KITTI}
\end{figure}

\begin{figure}[htbp]
\vspace{-0.2cm}
    \centering
    \includegraphics[scale=0.49]{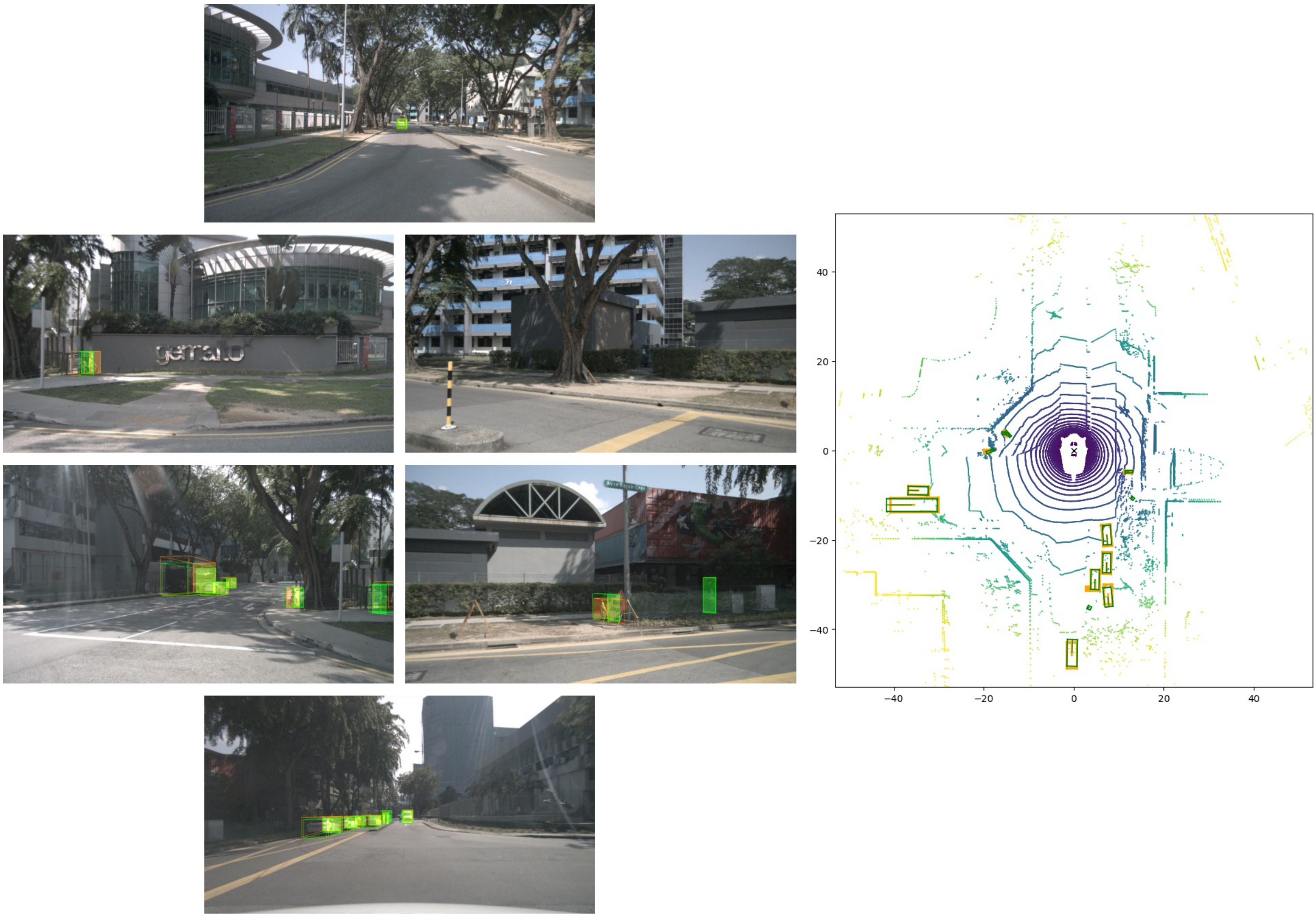}
    \vspace{-0.2cm}
    \caption{Visualization of the detection results of a sample from the nuScenes validation set. The estimation results of the model are in green and the ground truth boxes are marked in orange. The position of an image is set according to the corresponding camera position on the vehicle.} \label{Fig: Detection results on nuScenes}
\end{figure}

In this part, we visualize some detection results on KITTI-3D and nuScenes, which are illustrated in Fig. \ref{Fig: Detection results on KITTI} and Fig. \ref{Fig: Detection results on nuScenes}, respectively.

As shown in these two figures, the detection results are promising. Notably, there is an interesting failure case in Fig. \ref{Fig: Detection results on KITTI}, which is marked with a red ellipse. This failure is because a car is reflected by the smooth surface of a bus. Although this problem could result in incorrect decision making, it has not been studied in any publication.

\subsection{Where the Well-trained 2D Object Detectors Come from}
\label{Appendix: Where the well-trained 2D object detectors come from}

In the setting of all our experiments, we have two parts of data, the pre-training part and fine-tuning part. The images in the pre-training part are not labeled, but their corresponding lidar points are available. In the fine-tuning part, all images are annotated, which means their 3D bounding box labels are available. To train a 2D object detector that can label the images in the pre-training part with 2D boxes, we first convert the 3D bounding box labels of the images in the fine-tuning part as 2D boxes, and use these 2D boxes and the fine-tuning images to train a 2D object detector (CenterNet). In this way, a well-trained 2D object detector is obtained without extra labeling effort. Through visualization, we confirm that the quality of the generated 2D boxes by this 2D object detector on the fine-tuning images is promising.

If without a special statement, for KITTI, the pre-training part is KITTI-raw and the fine-tuning part is the KITTI-3D training set. In nuScenes, the pre-training part is the whole nuScenes training set and the fine-tuning part is the fixed 6000 samples among the nuScenes training set.

\subsection{More Discussion about this Work}
\label{Appendix: More Discussion about this Work}

In this work, we first point out that it is infeasible to design a M3OD pre-training task without any label information, because using only images to pre-train models (such as contrastive learning) can only produce image-level representation. By contrast, M3OD demands depth representation and instance-level appearance representation. Therefore, when we design M3OD pre-training tasks, what we should try to realize is only using labels that are easy to obtain, such as the 2D boxes and lidar points.

Another point we highlight is the guideline of designing pre-training tasks, i.e., making the pre-training imitate the target task representation. In this work, we employ two tasks (depth estimation and 2D object detection) to imitate M3OD.

Furthermore, we argue that pre-training is exactly a tool to exploit the information of extra data. Therefore, using extra data is important. If the pre-training data is the same as the fine-tuning data, only a small performance gain can be obtained.

\subsection{Source Code}
 
Refer to the supplemental material for the source code.

\end{document}